\documentclass[10pt,twocolumn,letterpaper]{article}

\usepackage{cvpr}              % To produce the CAMERA-READY version
\definecolor{cvprblue}{rgb}{0.21,0.49,0.74}
\usepackage[pagebackref,breaklinks,colorlinks,allcolors=cvprblue]{hyperref}
\usepackage{makecell}
\usepackage{algorithm}
\usepackage{algorithmic}
\usepackage{amsmath}
\usepackage{kotex}
\usepackage{amssymb}
\usepackage{adjustbox}
\usepackage{threeparttable}
\usepackage{booktabs}
\usepackage{siunitx}
\usepackage{multirow}
\usepackage{caption}
\usepackage[accsupp]{axessibility}  % Improves PDF readability for those with disabilities.
\usepackage{tabularx}
\usepackage[table]{xcolor}
\usepackage{hyperref}
\usepackage{xcolor}
\definecolor{cvprblue}{rgb}{0.21,0.49,0.74}
\usepackage{graphicx}
\def\paperID{19027} % *** Enter the Paper ID here
\def\confName{CVPR}
\def\confYear{2026}

\title{\textit{UniSpector}: Towards Universal Open-set Defect Recognition via Spectral-Contrastive Visual Prompting}
\author{
    Geonuk Kim \quad Minhoi Kim \quad Kangil Lee \quad Minsu Kim \\
    Hyeonseong Jeon \quad Jeonghoon Han \quad Hyoungjoon Lim \quad Junho Yim \\
    LG Energy Solution \\
}

\begin{document}
\maketitle
\begin{abstract}
Although industrial inspection systems should be capable of recognizing unprecedented defects, most existing approaches operate under a closed-set assumption, which prevents them from detecting novel anomalies. While visual prompting offers a scalable alternative for industrial inspection, existing methods often suffer from prompt embedding collapse due to high intra-class variance and subtle inter-class differences. To resolve this, we propose \textit{UniSpector}, which shifts the focus from naive prompt-to-region matching to the principled design of a semantically structured and transferable prompt topology. \textit{UniSpector} employs the Spatial-Spectral Prompt Encoder to extract orientation-invariant, fine-grained representations; these serve as a solid basis for the Contrastive Prompt Encoder to explicitly regularize the prompt space into a semantically organized angular manifold. Additionally, Prompt-guided Query Selection generates adaptive object queries aligned with the prompt. We introduce \textit{Inspect Anything}, the first benchmark for visual-prompt-based open-set defect localization, where \textit{UniSpector} significantly outperforms baselines by at least 19.7\% and 15.8\% in $\mathrm{AP}{50}^b$ and $\mathrm{AP}{50}^m$, respectively. These results show that our method enable a scalable, retraining-free inspection paradigm for continuously evolving industrial environments, while offering critical insights into the design of generic visual prompting. \href{https://geonuk-kimmm.github.io/UniSpector}{https://geonuk-kimmm.github.io/UniSpector}
\end{abstract}
    
\section{Introduction}
\label{sec:intro}
\begin{figure}[!t]
     \centering
     \includegraphics[width=\columnwidth,height=8.0cm]{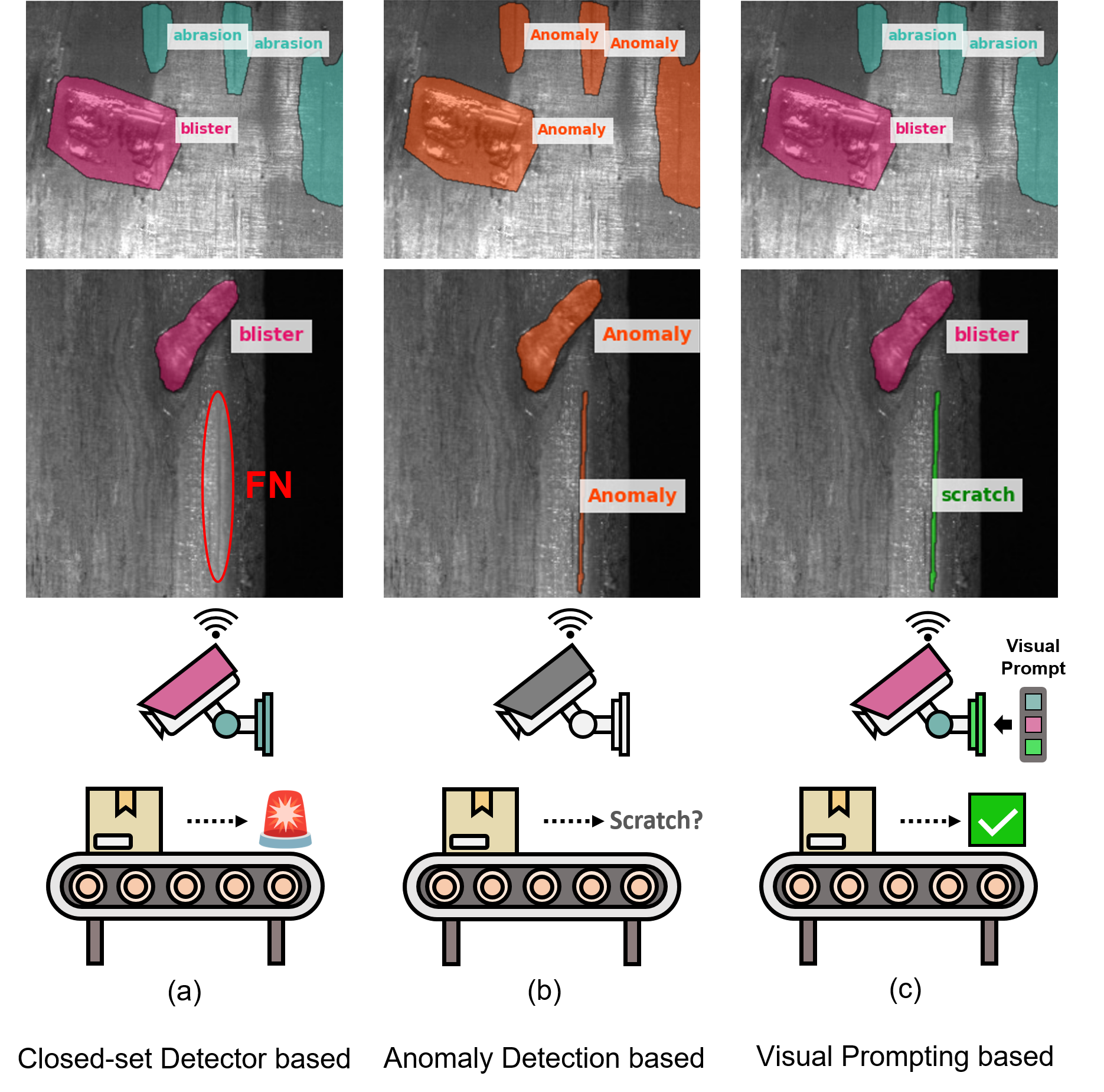}
     \caption{Comparison of visual inspection paradigms: (a) closed-set detectors fail on novel defect types and require costly retraining, (b) anomaly detectors cannot distinguish between defect classes, and (c) visual prompting enables open-set recognition by aligning unseen defects with exemplar prompts, providing a scalable visual inspection framework.}
     \label{fig:abstract}
     \vspace{-3mm}
\end{figure}

In industrial mass production, visual inspection is essential for quality control, where defect localization often matters more than merely declaring whether a product is defective. While existing supervised detectors perform well in closed-set scenarios with fixed defect categories~\cite{wang2023defect, liu2022adaptive, yan2025wavelet}, such assumptions rarely hold in practice, as new defect types continuously emerge and the definition of “normal” may shift over time. 
As illustrated in Figure~\ref{fig:abstract}, conventional closed-set models~\cite{yan2025wavelet,ding2024lerenet, liu2025exploring, hu2025category,ronneberger2015u} fail when new defects appear, requiring costly data rebuilding and retraining, whereas anomaly detection methods~\cite{roth2022towards,batzner2024efficientad,defard2021padim} cannot distinguish among defect types.

\begin{figure}[h]
     \centering
     \includegraphics[height=6cm]{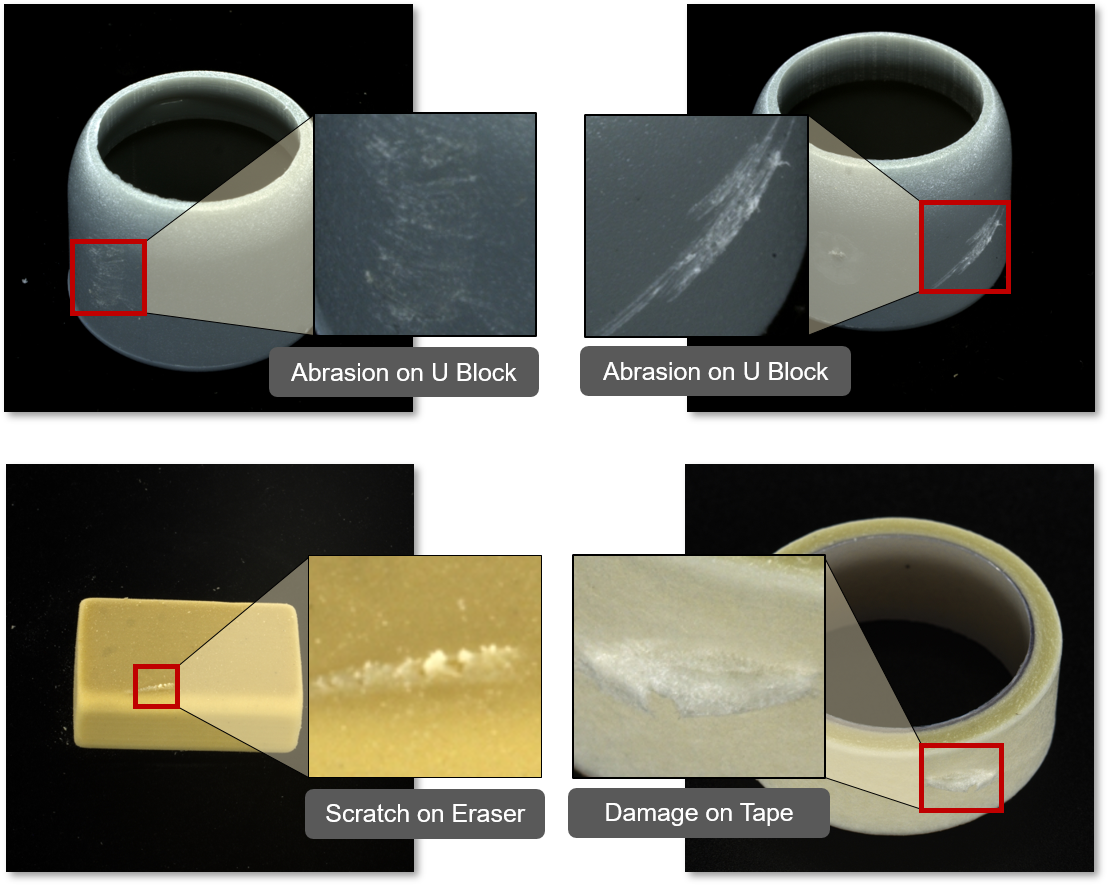}
     \caption{
Examples from the InsA benchmark. 
\textbf{Top:} Samples from the same defect class showing high intra-class appearance variance. 
\textbf{Bottom:} Samples from different classes exhibiting similar visual patterns, resulting in low inter-class separability. 
Such ambiguities highlight the inherent difficulty of defect recognition.
}
     \label{fig:similar_defects}
     \vspace{-3mm}
\end{figure}

To address category scalability without retraining, language-driven open-set detectors leverage text descriptions~\cite{liu2024grounding,jiang2024t,wang2025yoloe}. However, in fine-grained industrial inspection, defects are often visually subtle and hard to verbalize even for domain experts, making purely text-driven grounding unreliable. Consequently, visual prompting~\cite{li2024visual,zou2023segment, wang2023seggpt, liu2024simple}—which matches input image regions to exemplar prompts—emerges as a scalable alternative for open-set localization.
Nevertheless, we find that existing visual prompting methods suffer from prompt embedding collapse (Fig.~\ref{fig:prompt_embeddings}) when confronted with high intra-class variance and subtle inter-class differences typical of industrial domains (Fig.~\ref{fig:similar_defects}). By treating prompt embeddings merely as \textit{implicit representations learned via prompt-to-region matching}, prior works leave underexplored how the prompt embedding space should be structured to ensure robust transferability. We argue that applying visual prompting to the industrial domain reveals these critical but underexplored structural aspects of the visual prompting mechanism.

To overcome these limitations, we propose \textit{UniSpector}, a novel visual prompting framework for open-set defect detection and segmentation. Shifting the focus from prompt-to-region matching to the principled design of a semantically structured and transferable prompt topology. UniSpector comprises two core components: the Spatial–Spectral Prompt Encoder (SSPE) and the Contrastive Prompt Encoder (CPE).
Specifically, SSPE redefines the visual prompt as a dual-domain descriptor that captures both spatial and spectral features. It extracts orientation-invariant frequency cues to suppress intra-class variance caused by rotational transformations, while enhancing inter-class separation through fine-grained spatial-spectral representations capturing subtle textures often overlooked in the pixel domain. Building on these fine-grained representations, CPE explicitly regularizes the prompt embedding space to establish a semantically structured topology within an angular manifold through contrastive learning. By enforcing compactness within each class and sufficient angular separation between distinct defects, CPE ensures a semantically structured and transferable topology, where related defect types lie closer in angle and dissimilar ones farther apart—enabling semantically consistent placement of unseen defects instead of arbitrary projection. In addition, to improve adaptability to novel defect types, we introduce Prompt-guided Query Selection (PQS), which dynamically selects prompt-conditioned object queries via differentiable Gumbel–Softmax selection, ensuring that queries are adaptively initialized to the characteristics of prompt exemplar.

To evaluate the effectiveness, we establish Inspect Anything (InsA), the first benchmark for visual-prompt-based open-set defect detection. InsA measures both in-domain and cross-domain generalization ability, reflecting adaptability to real-world settings where visually similar defects may arise from different products or materials. 
Extensive experiments demonstrate that \textit{UniSpector} substantially outperforms existing baselines, achieving a 19.7\% higher detection $\mathrm{AP}_{50}^b$ and a 15.8\% higher segmentation $\mathrm{AP}_{50}^m$. These results highlight that the principled design of the prompt space to mitigate intra-class variance and enhance inter-class separability is key to open-set defect detection. Consequently, \textit{UniSpector} establishes a scalable foundation for next-generation inspection, eliminating the need for extensive labeling and retraining in evolving environments.

% This significant improvement highlights the practical effectiveness of visual prompting, establishing a reusable inspection paradigm capable of adapting to newly emerging defect types with minimal overhead.

\noindent\textbf{Contributions}
\begin{enumerate}
    \item Through the lens of industrial inspection,  we identify underexplored structural aspects of visual prompting, where treating prompts as implicit representations can lead to prompt embedding collapse due to high intra-class variance and subtle inter-class differences. Our findings offer key architectural insights for designing robust, transferable visual prompting mechanisms. 
    \item We propose \textit{UniSpector}, a novel framework where SSPE extracts orientation-invariant, dual-domain representations that serve as a solid basis for CPE to explicitly regularize the prompt space into a semantically structured angular manifold, ensuring enhanced discriminability and transferability for novel defects.
    \item We establish InsA, the first benchmark for visual-prompt-based open-set defect recognition. Experiment results show that \textit{UniSpector} surpasses baselines in both in-domain and cross-domain settings, demonstrating scalable, class- and domain-generalizable inspection.
\end{enumerate}

\begin{figure*}[!t]
 \begin{center}
 \includegraphics[height=9cm]{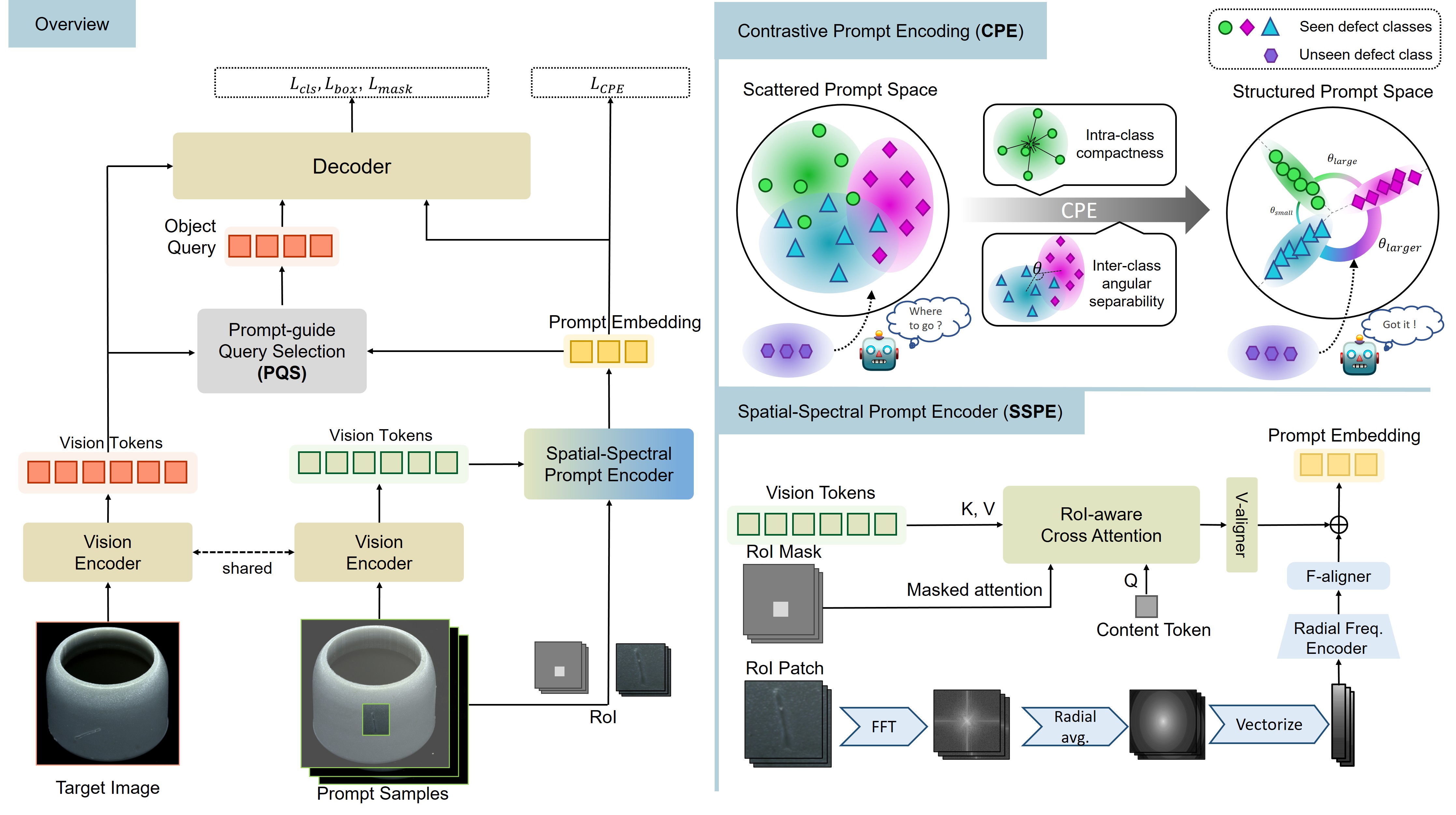}
 \end{center}
 \caption{Overview of \textit{UniSpector}, an open-set defect detection and segmentation framework. The Spatial–Spectral Prompt Encoder extracts orientation-invariant spectral cues fused with spatial features to distinguish visually similar defects. Building on these, Contrastive Prompt Encoding regularizes the prompt embedding space into a structured manifold for robust open-set generalization. A Prompt-guided Query Selection mechanism dynamically selects prompt-aware queries. Refer to Appendix~\protect\hyperlink{supp:section31}{C.1} for the inference phase architecture.}
\label{fig:architecture}
\end{figure*}

\section{Related Works}
\subsection{Traditional defect detection and segmentation}
Modern industrial defect inspection largely relies on deep learning models designed for instance- or pixel-level localization. 
U-Net–style segmentation networks~\cite{ronneberger2015u} remain widely adopted for their precise boundary modeling~\cite{liu2022adaptive, yang2022dual}, SSD-FasterNet~\cite{wang2022ssd} enhances Faster R-CNN with region proposals and FPN to capture fine-grained defects. 
Transformer-based architectures further improve long-range context modeling~\cite{wang2023defect,jiang2023cinformer}. 
WPA-QFormer~\cite{yan2025wavelet} uses wavelet decomposition and prototype queries to stabilize features for heterogeneous textures. 
Recent studies~\cite{ding2024lerenet, liu2025exploring, hu2025category} have explored strategies to mitigate high intra-class variability and low inter-class separability due to variations in texture, illumination, or geometry.
Despite strong supervised performance, all of these methods assume a fixed defect taxonomy and operate in a closed-set regime, making them unsuitable for real-world industrial environments where novel defect types continue to emerge.

\subsection{Visual/Text-prompted Open-Set Detection}
GroundingDINO~\cite{liu2024grounding} and YOLO-World~\cite{cheng2024yolo} introduce novel object detection by associating language descriptions with visual features, without relying on pre-defined categories. However, in domains involving specialized or technical objects, verbal descriptions can be challenging. DINOv~\cite{li2024visual} explores the use of visual prompts as in‑context examples for general vision tasks, while T-Rex2~\cite{jiang2024t} and YOLOE~\cite{wang2025yoloe} further integrate both text and visual prompts. COSINE~\cite{liu2025unified} leverages multiple backbones to extract multi-modal features, and Prompt-DINO~\cite{guan2025text} applies order-aware query selection to align text and visual cues. Despite these advances, most methods assume well-defined classes—assumptions rarely met in industrial defect inspection. Their focus on prompt-to-region matching neglects structure of the prompt embedding space, causing semantic collapse when encountering novel or fine-grained defects.

\section{Methods}
We propose \textit{UniSpector}, a visual prompting framework tailored for open-set defect detection and segmentation. To mitigate prompt embedding collapse inherent in prompt-to-region matching due to high intra-class variance and subtle inter-class difference, \textit{UniSpector} employs two critical components:  
(1) the Spatial-Spectral Prompt Encoder (SSPE), which enhances inter-class separability by integrating fine-grained spectral descriptors to distinguish subtle textures beyond the pixel domain, while suppressing intra-class variance via orientation-invariant radial frequency cues to handle geometric rotations; and (2) leveraging these refined representations, the Contrastive Prompt Encoder (CPE) explicitly regularizes the embedding manifold to establish a semantically organized angular topology, ensuring intra-class compactness and sufficient inter-class separation. Furthermore, Prompt-guided Query Selection (PQS) dynamically
selects prompt-conditioned object queries.

\subsection{Overview}
As shown in Fig.~\ref{fig:architecture}, \textit{UniSpector} comprises the DETR framework. The vision encoder extracts token embeddings $\mathcal{F} \in \mathbb{R}^{N_s \times d}$ from a target image, where $N_s$ and $d$ denote the number of tokens and the hidden dimension, respectively. 
Given $K$ annotated prompt samples, prompt embeddings are averaged per class to obtain class-level prototypes $\mathbf{p}_c \in \mathbb{R}^d$, where $c$ is the class index. To obtain $\mathcal{F}$-conditioned representations, the decoder performs cross-attention $\textit{CA}(\cdot)$ over $\mathcal{F}$ using object queries $\mathbf{q} \in \mathbb{R}^{N_q \times d}$ and prototypes $\mathbf{p}_c$, followed by a linear projection $h(\cdot)$. Here, $N_q$ denotes the number of object queries. This process yields:
\begin{equation}
\hat{\mathbf{q}}_i = h(\textit{CA}(\mathbf{q}_i, \mathcal{F}))  \text{,} \quad \hat{\mathbf{p}}_c = h(\textit{CA}(\mathbf{p}_c, \mathcal{F}))
\end{equation}
The final logit $y_{i,c}$ is then computed via the dot product between the class-level prototypes and query embeddings:
\begin{equation}
y_{i,c} = (\hat{\mathbf{p}}_c)^\top \hat{\mathbf{q}}_i, \quad 
i = 1,\dots,N_q, \; c = 1,\dots,C
\end{equation}
%\begin{equation}
%y_{i,c} = \frac{(\hat{\mathbf{p}}^*_c)^\top \hat{\mathbf{q}}_i}{\|\hat{\mathbf{p}}^*_c\|_2 \, \|\hat{\mathbf{q}}_i\|_2}, \quad 
%i = 1,\dots,N_q, \; c = 1,\dots,C
%\end{equation}
where $C$ denotes the number of unique classes among the $K$ prompt samples. This design allows flexible class-conditioned predictions without relying on a fixed classifier, supporting robust recognition of unseen defect categories. The encoder-decoder architecture and the training-inference phase are detailed in Appendices~\protect\hyperlink{supp:section31}{C.1} and \protect\hyperlink{supp:section33}{C.3}, respectively. %Encoder-decoder and training-inference disentanglement details are in Appendix~\suppref{3.1} and \suppref{3.3}, respectively.

\subsection{Spatial-Spectral Prompt Encoder}
For each of the $K$ annotated prompt samples, we extract the corresponding RoI patch $R_k$ based on the annotated bounding box. To capture orientation-invariant yet fine-grained features, we first compute the 2D Discrete Fourier Transform (DFT) of each patch, $F_k(u,v) = \textit{DFT}(R_k)$. We then eliminate directional bias by averaging the magnitude responses along concentric frequency rings:
\begin{equation}
h_k(\rho) = \frac{1}{|\Gamma_\rho|}\sum_{(u,v)\in\Gamma_\rho} \bigl|F_k(u,v)\bigr|,
\end{equation}
where $\Gamma_\rho$ denotes the set of frequency components at a radial distance $\rho$ from the origin. The resulting 1D radial spectrum $h_k$ preserves global structural and spectral patterns while remaining orientation-invariant.

To derive the final multi-modal representation, we employ a dual-path encoding strategy. First, the spectral code $z_k^{\text{freq}}$ is obtained by projecting $h_k$ into a latent space via a Radial Frequency Encoder. In parallel, a spatial code $z_k^{\text{spatial}}$ is extracted from the prompt image using masked cross-attention. This attention mechanism is guided by the RoI mask of the prompted region, ensuring that only the defect-relevant area contributes to the representation. To effectively fuse these modalities, we employ two lightweight MLPs, denoted as $f_{\text{align}}$ and $v_{\text{align}}$. The final prompt embedding $\mathbf{e}$ for the $k$-th sample is then computed as:
\begin{equation}
{\mathbf{e}}_k = f_{\text{align}}(z_k^{\text{spatial}}) + v_{\text{align}}(z_k^{\text{freq}})
\end{equation}
Through this dual-path fusion, \textit{UniSpector} enhances inter-class separability even under high visual similarity, while the orientation-invariant radial spectrum effectively minimizes intra-class variance caused by rotations. Detailed derivations of the radial frequency extraction are provided in the Appendix~\protect\hyperlink{supp:section2}{B}.

\subsection{Contrastive Prompt Encoding}
\label{sec:cpe}
We introduce the CPE, which explicitly regularizes the prompt embedding space to exhibit semantically coherent topology to enhance open-set transferability. Specifically, CPE encourages two complementary properties:  
(1) \emph{Intra-class compactness}: Aligning instances of the same class into a shared semantic identity, ensuring invariance to geometric variations (e.g., orientation, width).
(2) \emph{Semantic angular topology}: Structuring the embedding manifold such that semantically related classes occupy neighboring regions while dissimilar ones are separated by wider angular margins.
To encourage this structure during training, we employ the prompt embeddings with an angular-margin contrastive loss, inspired by ArcFace~\cite{deng2019arcface}.  
For each defect class \(c\), a class-level prototype \(\mathbf{p}_c\) is computed by averaging the embeddings $\mathbf{e}$ of same-class instances among the $K$ samples.  
The cosine similarity between a prompt embedding \(\mathbf{e}_k\) and its class prototype \(\mathbf{p}_c\) is then given by
\begin{equation}
\cos \theta_{c,k} = 
\frac{(\mathbf{p}_c)^\top \mathbf{e}_k}
{\lVert \mathbf{p}_c \rVert_2 \, \lVert \mathbf{e}_k \rVert_2 }
\end{equation}
We then apply an angular-margin loss with margin $m$:
\begin{equation}
\resizebox{\columnwidth}{!}{$
\mathcal{L}_{\text{CPE}} =
-\frac{1}{N}\sum_{k=1}^{N}
\log
\frac{\exp\big(\alpha \cdot \cos(\theta_{y_k,k}+m)\big)}
{\exp\big(\alpha \cdot \cos(\theta_{y_k,k}+m)\big)
+ \sum_{c \ne y_k} \exp\big(\alpha \cdot \cos(\theta_{c,k})\big)}
$}
\end{equation}
where $\alpha$ is a scaling factor, $y_k$ is the ground-truth class label of the $k$-th instance, and $N$ is the number of prompt instances in the batch.
The margin $m$ enforces a minimum angular gap between classes, preventing embedding collapse. Importantly, once this margin is satisfied, embeddings are free to self-organize according to semantic affinity—allowing related classes to stay closer while pushing dissimilar classes farther apart, thus yielding a structured and semantically meaningful angular space. Consequently, this structured topology enables semantically consistent placement of unseen defects instead of arbitrary projection. For the full objective function integrating $\mathcal{L}_{\text{CPE}}$ and further details, please refer to Appendix~\protect\hyperlink{supp:section32}{C.2}.

\subsection{Prompt-guided Query Selection}

Conventional DETR-style detectors and recent visual prompting approaches~\cite{li2024visual,liu2025unified} rely on learnable object queries that encode biases from learned classes, limiting open-set generalization. Although similarity-based top-$K$ heuristics~\cite{liu2024grounding,jiang2024t} have been explored, they are non-differentiable and cannot be optimized end-to-end.

To address this, we introduce a prompt-guided query selection module that selects a subset of vision tokens as object queries, conditioned on the given visual prompt. Given the vision token embeddings $\mathcal{F} \in \mathbb{R}^{N_s \times d}$ and class prototypes $\mathbf{p} \in \mathbb{R}^{C \times d}$, we first compute cosine similarities between each token and all prompts, then reduce them to a single relevance score per token by taking the maximum over class prototypes, resulting in a vector $\mathbf{S} \in \mathbb{R}^{N_s}$. To enable differentiable top-$K$ selection, we add element-wise Gumbel noise $\mathbf{g} \in \mathbb{R}^{N_s}$ and compute soft selection weights:
\[
\tilde{\mathbf{w}} = \mathrm{softmax}\big((\mathbf{S} + \mathbf{g}) / \tau\big),
\]
where $\tau$ is a temperature parameter. The hard top-$K$ tokens are then selected based on $\tilde{\mathbf{w}}$ to form a binary mask, which is used in the forward pass while gradients flow through $\tilde{\mathbf{w}}$ via a Straight-Through Estimator. This produces prompt-conditioned object queries from the vision tokens in a fully differentiable manner, eliminating reliance on static priors.

\section{Experiments}
Industrial defects are typically small, subtle, and visually heterogeneous, often lacking consistent semantics or clear category boundaries. To evaluate generalization ability to unseen defects, we introduce Inspect Anything (InsA), a benchmark for open-set defect detection and segmentation within a visual prompting framework.

InsA evaluates three aspects of generalization:
(1) in-domain generalization to novel defect types that share visual similarity with the seen sets,
(2) cross-domain transferability to unseen defects emerging under different material properties, imaging conditions, and defect morphologies, and
(3) prompt-level generalization, quantified by aggregating results across multiple class partitions and independently sampled prompt sets to reflect stability rather than dependence on a single prompt configuration.

\begin{table*}[h]
\small
  \centering
  \begin{center}
  \begin{tabular}{c|ccc|cc|c}
    \hline
    \multirow{3}{*}{\textbf{Dataset}} & \multicolumn{3}{c|}{\textbf{Count}} & \multicolumn{3}{c}{\textbf{Class Split}} \\
    \cline{2-7}
     & \multirow{2}{*}{Images} & \multirow{2}{*}{Defect Instances} & \multirow{2}{*}{Defect Categories} & \multicolumn{2}{c|}{\textit{In-domain}} & \multirow{2}{*}{\textit{Cross-domain}} \\
    \cline{5-6}
     &  &  &  & Seen (Train) & Unseen (Test) &  \\
    \hline % \midrule
    
    GC10 ~\cite{lv2020deep}                 & 2,292 & 3,563 & 10 & 7 & 3 & - \\
    MagneticTile ~\cite{huang2020surface}   & 388   & 514   & 5 & 4 & 1 & - \\
    Real-IAD ~\cite{wang2024real}           & 49,237 & 55,326 & 111 & 83 & 28 & - \\
    MVTec ~\cite{bergmann2019mvtec}         & 1,258 & 1,836 & 73 & 53 & 20 & - \\
    3CAD ~\cite{yang20253cad}               & 11,074 & 16,559 & 46 & - & - & 46 \\
    VISION ~\cite{bai2023vision}            & 1,757 & 3,553 & 44 & - & - & 44 \\
    VisA ~\cite{zou2022spot}                & 1,167 & 2,131 & 71 & - & - & 71 \\
    \midrule
    
    \textbf{Total} & 67,173 & 83,482 & 360 & 147 & 52 & 161 \\
    \bottomrule
  \end{tabular}
  \end{center}
  \caption{Overview of the InsA benchmark composition. The \textit{in-domain} split contains seen sets for training and unseen sets held out for open-set evaluation, while the \textit{cross-domain} split is held out for cross-domain evaluation. Refer to the Appendix~\protect\hyperlink{supp:section1}{A} for the details.}
  \label{tab:insa-benchmark}
\end{table*}

\begin{table*}[t]
  \centering
  \setlength{\tabcolsep}{1.6mm} % 열 간격 줄이기
  \resizebox{\textwidth}{!}{% 폭을 전체 페이지에 맞춤
  \begin{tabular}{l|cccccccc|cccccc|cc}
    \hline
    \multirow{3}{*}{Methods} & \multicolumn{8}{c|}{\textit{In-domain}} & \multicolumn{6}{c|}{\textit{Cross-domain}} & \multicolumn{2}{c}{\textit{Overall}}\\
    % \cline{2-17} % \cmidrule(lr){2-9} \cmidrule(lr){10-15} \cmidrule(lr){16-17}
    \cmidrule{2-17}
    & \multicolumn{2}{c}{GC10} & \multicolumn{2}{c}{MagneticTile} & \multicolumn{2}{c}{Real-IAD} & \multicolumn{2}{c|}{MVTec AD} 
    & \multicolumn{2}{c}{3CAD} & \multicolumn{2}{c}{VISION} & \multicolumn{2}{c|}{VisA} & \multicolumn{2}{c}{\textit{Avg}} \\
    & $\mathrm{AP}^{b}_{50}$ & $\mathrm{AP}^{m}_{50}$ & $\mathrm{AP}^{b}_{50}$ & $\mathrm{AP}^{m}_{50}$ & $\mathrm{AP}^{b}_{50}$ & $\mathrm{AP}^{m}_{50}$ & $\mathrm{AP}^{b}_{50}$ & $\mathrm{AP}^{m}_{50}$ & $\mathrm{AP}^{b}_{50}$ & $\mathrm{AP}^{m}_{50}$ & $\mathrm{AP}^{b}_{50}$ & $\mathrm{AP}^{m}_{50}$ & $\mathrm{AP}^{b}_{50}$ & $\mathrm{AP}^{m}_{50}$ & $\mathrm{AP}^{b}_{50}$ & $\mathrm{AP}^{m}_{50}$\\
    \midrule

    \textit{\small visual grounding} & & & & & & & & & & & & & & & & \\
    GroundingDINO~\cite{liu2024grounding} & 9.6 & - & 26.7 & - & 0.3 & - & 1.4 & - & 0.0 & - & 0.0 & - & 0.0 & - & 5.4 & - \\
    GroundingDINO$^\dagger$\cite{liu2024grounding} & 8.0 & - & 11.0 & - & 1.2 & - & 18.8 & - & 0.0 & - & 0.0 & - & 0.1 & - & 5.6 & - \\
    YOLO-World~\cite{cheng2024yolo} & 7.0 & 2.1 & 30.7 & 28.6 & 0.3 & 0.2 & 2.7 & 2.4 & 0.0 & 0.0 & 0.3 & 0.3 & 1.2 & 1.2 & 6.0 & 5.0 \\
    YOLO-World$^\dagger$\cite{cheng2024yolo} & 3.4 & 1.5 & 29.8 & 27.5 & 21.7 & 16.7 & 5.4 & 4.7 & 0.0 & 0.0 & 0.0 & 0.0 & 1.7 & 1.6 & 8.9 & 7.4\\
    \midrule
    \textit{\small visual prompting} & & & & & & & & & & & & & & & & \\
    SEEM~\cite{zou2023segment} & - & 0.2 & - & 0.2 & - & 0.0 & - & 0.3 & - & 0.0 & - & 0.0 & - & 0.0 & - & 0.1 \\
    SEEM$^\dagger$\cite{zou2023segment} & - & 0.1 & - & 0.6 & - & 0.0 & - & 0.1 & - & 0.0 & - & 0.0 & - & 0.0 & - & 0.1 \\
    SegGPT~\cite{wang2023seggpt} & - & 16.0 & - & 20.1 & - & 3.1 & - & 6.8 & - & 1.7 & - & 0.9 & - & 1.7 & - & 7.2 \\
    SINE~\cite{liu2024simple} & 0.7 & 0.5 & 2.0 & 2.0 & 0.6 & 0.5 & 4.6 & 4.4 & 0.1 & 0.2 & 0.7 & 0.6 & 0.9 & 0.9 & 1.4 & 1.3 \\
    SINE$^\dagger$\cite{liu2024simple} & 0.9 & 0.9 & 1.6 & 1.9 & 1.2 & 0.5 & 4.5 & 4.1 & 0.1 & 0.1 & 0.6 & 0.2 & 1.0 & 0.8 & 1.4 & 1.2 \\
    DINOv~\cite{li2024visual} & 3.2 & 0.8 & 30.0 & 26.9 & 2.2 & 1.4 & 19.0 & 15.0 & 4.1 & 2.1 & 4.3 & 3.8 & 8.4 & 7.0 & 10.2 & 8.1\\
    DINOv$^\dagger$\cite{li2024visual} & 16.5 & 16.6 & 48.4 & 39.6 & 21.0 & 17.5 & 15.9 & 15.2 & 2.9 & 1.9 & 4.6 & 3.8 & 10.4 & 8.5 & 17.1 & 14.7\\
    T-Rex2 $^\dagger$~\cite{jiang2024t} & 32.4 & 33.9 & 49.0 & 38.0 & 25.1 & 28.8 & 24.4 & 22.4 & 4.3 & 2.9 & 5.4 & 4.3 & 7.8 & 6.7 & 21.2 & 19.6 \\
    YOLOE~\cite{wang2025yoloe} & 1.6 & 0.4 & 48.3 & 45.4 & 16.6 & 13.9 & 26.9 & 22.7 & 4.9 & 2.0 & 7.2 & 5.9 & 14.4 & 12.2 & 17.1 & 14.6\\
    YOLOE$^\dagger$\cite{wang2025yoloe} & 10.7 & 9.5 & 43.3 & 41.8 & 17.2 & 15.5 & 25.8 & 23.9 & 3.3 & 1.4 & 3.5 & 3.0 & 17.7 & 15.3 & 17.4 & 15.8\\
    \rowcolor{blue!7}UniSpector (Ours)$^\dagger$ & \textbf{38.2} & \textbf{36.9} & \textbf{63.3} & \textbf{57.7} & \textbf{69.1} & \textbf{56.7} & \textbf{53.5} & \textbf{46.5} & \textbf{14.1} & \textbf{10.0} & \textbf{15.3} & \textbf{12.5} & \textbf{32.8} & \textbf{27.8} & \textbf{40.9} & \textbf{35.4}\\
    \bottomrule
  \end{tabular}}
  \caption{Open-set detection and segmentation performance on the InsA. $^{\dagger}$  Denotes models fine-tuned on in-domain seen sets of InsA.}
  \label{tab:main}
\end{table*}

\begin{table*}[t]
\begin{center}
\centering
\caption{Ablation study on individual modules in \textit{UniSpector}. All experiments were conducted on the Real-IAD test set. }
\label{tab:ablation_main}
\begin{threeparttable}
\small
\begin{tabular}{c|c|c|cccccc|cccccc}
\hline
\multirow{2}{*}{SSPE} & \multirow{2}{*}{CPE} & \multirow{2}{*}{PQS} 
&  \multicolumn{6}{c|}{Detection} & \multicolumn{6}{c}{Segmentation} \\ \cline{4-15}
 & & & $\mathrm{AP}^{b}$ & $\mathrm{AP}^{b}_{50}$& $\mathrm{AP}^{b}_{75}$ & $\mathrm{AP}^{b}_{s}$ &$\mathrm{AP}^{b}_{m}$ &$\mathrm{AP}^{b}_{l}$ & $\mathrm{AP}^{m}$ & $\mathrm{AP}^{m}_{50}$ & $\mathrm{AP}^{m}_{75}$ & $\mathrm{AP}^{m}_{s}$ &$\mathrm{AP}^{m}_{m}$ & $\mathrm{AP}^{m}_{l}$ \\ \hline
 &  &  & 13.6  & 24.0 & 14.5  & 17.9 & 16.3  & 24.0  & 7.7 & 20.0 & 5.1 & 7.2 & 12.3 & 29.6 \\
 {\checkmark} &  &  & 27.9 & 43.0 & 31.0  & 38.3 & 31.4  & 39.8  & 17.7 & 34.8 & 16.9 & 17.2 & 24.8 & 33.7 \\
 {\checkmark} & {\checkmark}  &  & 43.8  & 65.8 & 48.9  & 47.4& 47.6  & 44.7 & 26.0 & 53.1 & 22.3 & 21.0 & 32.7 & 36.2 \\
{\checkmark} & {\checkmark} & {\checkmark}&   46.3  & 69.1 & 51.9  & 50.3& 50.1  & 47.8 & 28.9 & 56.7 & 24.5 & 23.3 & 34.1 & 38.0 \\ \hline
\end{tabular}
\end{threeparttable}
\end{center}
\end{table*}

\begin{table}[h]
\centering
\caption{Comparison of query selection strategies.}
\label{tab:ablation_PQS_multi}
\small
\begin{tabular}{c|cc|cc}
\hline
\multirow{2}{*}{Strategy} & \multicolumn{2}{c|}{GC10} & \multicolumn{2}{c}{VisA} \\ 
\cline{2-5}
                          & $\mathrm{AP}^{b}_{50}$ & $\mathrm{AP}^{m}_{50}$ & $\mathrm{AP}^{b}_{50}$ & $\mathrm{AP}^{m}_{50}$ \\ \hline
Learnable Parameter        & 34.4 & 34.2 & 29.2 & 24.7 \\ 
Heuristic Top-$K$          & 35.6 & 33.9 & 30.1 & 25.4 \\ 
PQS                        & 38.2 & 36.9 & 32.8 & 27.8 \\ \hline
\end{tabular}
\end{table}

\begin{table}[t]
\centering
\caption{In-domain evaluation on seen defect set using all categories of Real-IAD.
Unlike the main open-set benchmark, here we disable the seen/unseen split and train all models on the full label space, using a standard train/test split.}
\label{tab:closed_result}
\begin{adjustbox}{max width=\textwidth,center}
    \small
    \begin{tabular}{l|cc} 
        \toprule
        \multirow{2}{*}{\textbf{Method}} 
        & \multicolumn{2}{c}{\textbf{Real-IAD}} \\ 
        & $\mathrm{AP}^{b}_{50}$ & $\mathrm{AP}^{m}_{50}$ \\ \hline
        \multicolumn{3}{l}{\textit{\scriptsize Closed-set models}} \\
        YOLOv11~\cite{khanam2024yolov11}  & 88.3 & -    \\
        RT-DETR~\cite{zhao2024detrs} & 89.9 & - \\
        Mask2Former~\cite{cheng2022masked} & 88.8 & 82.0 \\
        MaskDINO~\cite{li2023mask} & 91.7 & 84.5  \\ \hline 
        \multicolumn{3}{l}{\textit{\scriptsize Visual-Prompting models}} \\
        DINOv~\cite{li2024visual} & 24.8 & 20.4 \\
        T-Rex2~\cite{jiang2024t} & 29.8 & 26.4 \\
        UniSpector (Ours)  & \textbf{90.0} & \textbf{82.7} \\
        \bottomrule
    \end{tabular}
\end{adjustbox}
\end{table}
\subsection{InsA Benchmark Overview}
\subsubsection{Dataset Composition}
As shown in Tab.~\ref{tab:insa-benchmark}, InsA aggregates seven publicly available industrial defect datasets. We construct an \emph{in-domain open-set} protocol by splitting four datasets (GC10, MagneticTile, Real-IAD, MVTec) into disjoint \emph{seen} (train) and \emph{unseen} (test) sets. To reduce sampling bias and ensure reproducibility, we generate three splits with fixed random seeds \(\{42, 82, 777\}\). All training samples for the seen sets across these four datasets are merged into a single training set. To assess cross-domain transfer, we further evaluate \emph{cross-domain} data drawn from 3CAD, VISION, and VisA, which contain imaging conditions and defect patterns that differ significantly from those seen during training. Unlike the in-domain, the cross-domain setting evaluates on \emph{all} classes of each dataset, measuring the model’s ability to transfer beyond domain boundaries as well. Detailed dataset information is provided in the Appendix~\protect\hyperlink{supp:section1}{A}.
\subsubsection{Evaluation}
We adopt an exemplar-based evaluation protocol. For each prompt seed 
\(s \in \{42, 82, 777\}\), we sample \(5\) exemplar images per class as prompts and use the remaining images as queries, without overlap between the two sets. 
This prompt-level seeding is independent of the class-split seeds that define the seen/unseen partitions.

\noindent\textbf{Metrics.}
We report \(\mathrm{AP}_{50}^{b}\) and \(\mathrm{AP}_{50}^{m}\) as the primary metrics as industrial defects often exhibit ambiguous or fuzzy spatial boundaries (e.g., dents, bends, scratches), 
making the precise extent of a ground-truth region inherently subjective. 
Thus, stricter IoU thresholds (e.g., 0.75 or 0.95) provide limited practical value.

\noindent\textbf{Reporting.}
We fine-tune each method on the \emph{seen} set under three class-split seeds 
\(\{42, 82, 777\}\), yielding three independently trained models.
Each model is evaluated under the three prompt–query partitions defined by the same seeds, 
resulting in \(3 \text{ (models)} \times 3 \text{ (prompt seeds)} = 9\) runs per setting. 
We report the mean across these runs for both in-domain unseen and cross-domain evaluations.
All baselines are compared under two configurations:  
(i) using publicly released pre-trained weights when available, and  
(ii) after fine-tuning on the InsA in-domain seen sets.
\subsection{Baseline Models}
We compare our approach against two categories of related methods: visual grounding models and visual prompting models. For the visual grounding baselines, GroundingDINO~\cite{liu2024grounding} and YOLO-World~\cite{cheng2024yolo} are trained using category names converted into caption-style prompts with a unified template, “a $\{defect\_name\}$ defect of the $\{product\_name\}$.” GroundingDINO is evaluated solely on detection as it has no segmentation head. SEEM~\cite{zou2023segment} is trained with both image and text prompts using the same caption template, and is evaluated solely on segmentation as it lacks a detection head. For the visual prompting baselines, SegGPT~\cite{wang2023seggpt} is evaluated using only the publicly released pre-trained weights, as no official training code is available. Consequently, it is not fine-tuned on InsA, and evaluation is performed at its supported input resolution of 448×448. All remaining models are evaluated under both pre-trained and benchmark fine-tuning settings for fair comparison. SINE~\cite{liu2024simple} and DINOv~\cite{li2024visual} are trained following their official visual prompting configurations without modification. 
T-Rex2~\cite{jiang2024t} is re-implemented (no official code is provided) and adapted to operate in purely visual prompting mode. 
YOLOE~\cite{wang2025yoloe} is evaluated with its textual interface disabled and modified to accept only visual prompts when fine-tuned on InsA. Implementation details are provided in the Appendix~\protect\hyperlink{supp:section31}{C.1,C.4}.

\begin{figure*}[h]
 \begin{center}
 \includegraphics[trim=20mm 20mm 20mm 20mm, clip, width=0.85\linewidth,height=9.5cm]{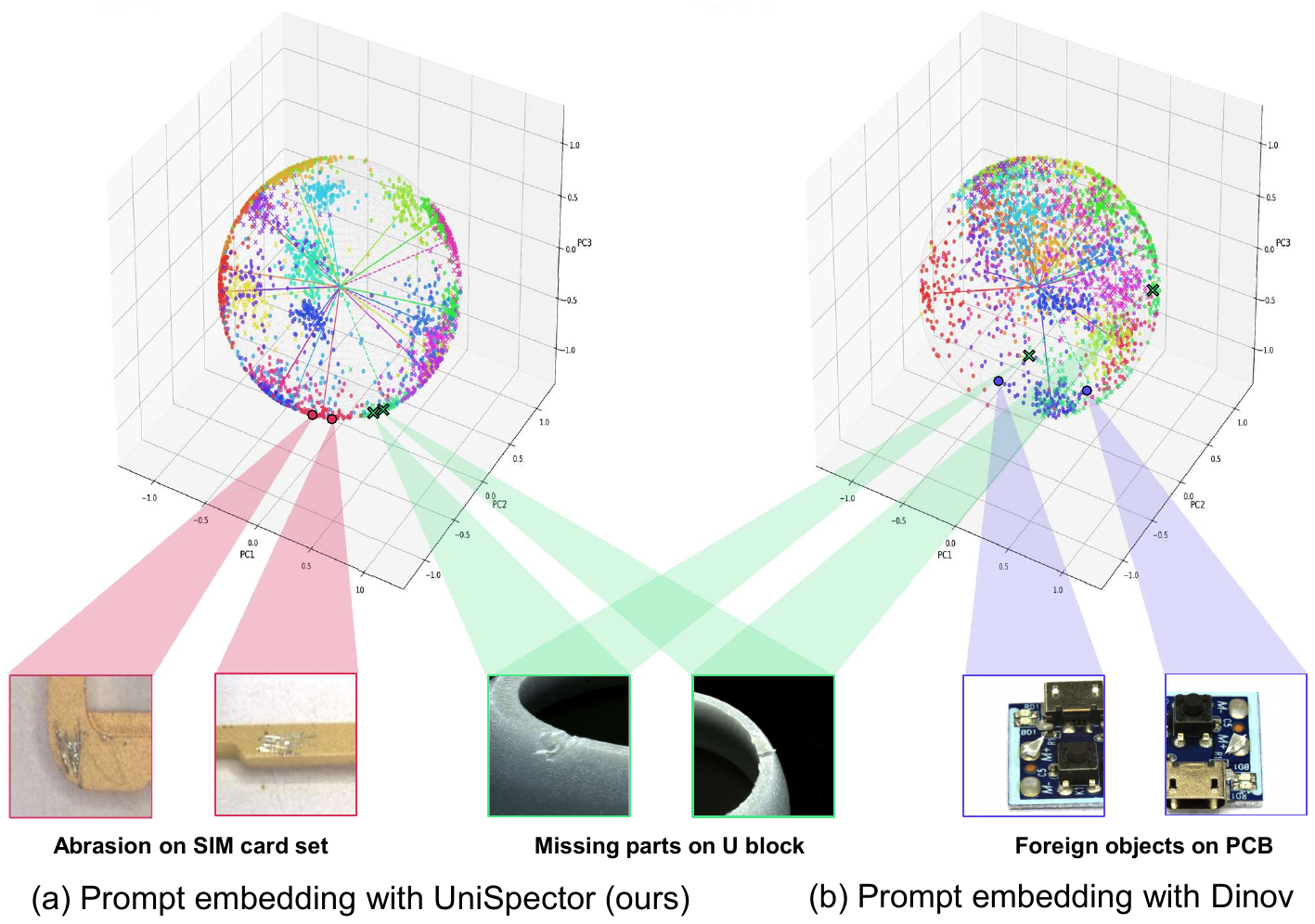}
 \end{center}
\caption{3D PCA projection of L2-normalized prompt embeddings learned by \textit{UniSpector} (left) and DINOv (right). Circle markers ($\circ$) denote seen sets and cross markers ($\times$) denote unseen sets.}
\label{fig:prompt_embeddings}
\end{figure*}

\subsection{Comparison on InsA Benchmark}
\noindent\textbf{Text-prompt for open-set inspection}
Tab.~\ref{tab:main} summarizes the performance of \textit{UniSpector} with a range of baseline methods. Visual grounding models such as GroundingDINO and YOLO-World exhibit modest performance on certain datasets, but their overall $\mathrm{AP}_{50}$ remains below 10\%, as many industrial defects cannot be fully described correctly by textual prompts. Notably, these visual grounding approaches show near-zero performance in cross-domain scenarios. These can be attributed to overfitting on highly specific textual labels corresponding to seen product defects, which limits their transferability compared to purely visual prompting methods.

\noindent\textbf{Visual-prompt for open-set inspection}  
Other baselines (SEEM, SegGPT, and SINE) underperform even after fine-tuning on the training set. Recent visual prompting methods, DINOv and YOLOE, achieve modest performance with a pre-trained model. Notably, despite DINOv being trained on over one billion images in SA-1B, the subtle and heterogeneous nature of industrial defects limits effective transfer from such large-scale pre-training. After fine-tuning on InsA, both DINOv and YOLOE show slight improvements, with $\mathrm{AP}^{b}_{50}$ increasing from 10.2\% to 17.1\% and from 17.1\% to 17.4\%, respectively. 

In contrast, \textit{UniSpector} consistently outperforms all baselines across both in-domain and cross-domain test sets by a substantial margin. Specifically, on \textit{MagneticTile} and \textit{Real-IAD}, \textit{UniSpector} achieves $\mathrm{AP}^{b}_{50}$ scores of 63.3\% and 69.1\%, respectively, demonstrating its strong potential to develop a visual inspection system that generalizes across diverse defect types. However, the cross-domain performance remains relatively lower than the in-domain results, primarily due to the strong dataset-specific imaging biases inherent in industrial inspection data. Such biases—arising from variations in illumination, texture, and capture conditions—pose a major challenge for generalization. While \textit{UniSpector} effectively learns transferable semantic structures, its sensitivity to these domain-specific variations indicates that enhancing domain robustness—through cross-domain regularization or domain-agnostic prompt representation—remains a promising direction for future work.

\subsection{Model Analysis}
\textbf{Effectiveness of each module}  
Ablation studies on Real-IAD (Tab.~\ref{tab:ablation_main}) confirm that each component incrementally improves performance. Specifically, given the refined prompt embeddings provided by SSPE, CPE yields the most significant gains by structuring the embedding space for open-set transfer. Furthermore, PQS improves objectness estimation through prompt-adaptive queries, with the full integration achieving the best results.

\begin{figure}
\begin{center}
 \includegraphics[width=1.0\linewidth,height=3cm]{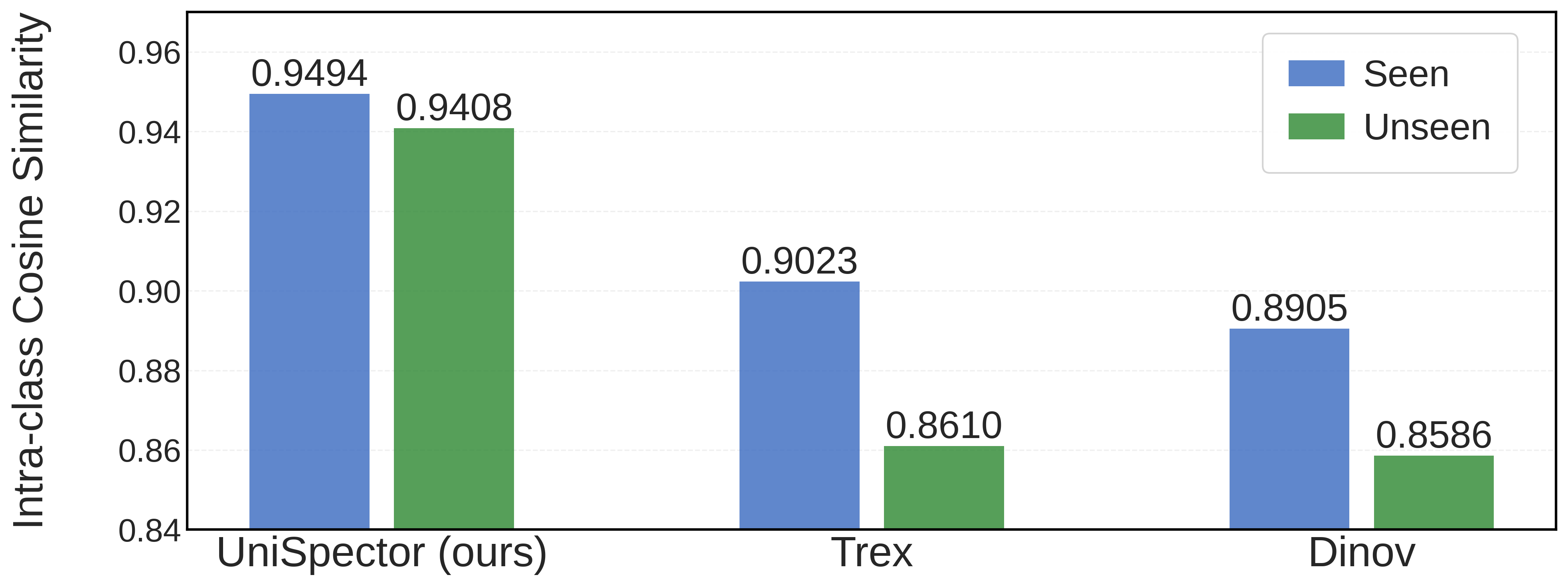}
 \end{center}
 \caption{Intra-class cosine similarity comparison across seen and unseen defect classes on RealIAD.}
\label{fig:intra-cosine}
\end{figure}

\begin{figure}
\begin{center}
 \includegraphics[width=1.0\linewidth,height=3.0cm]{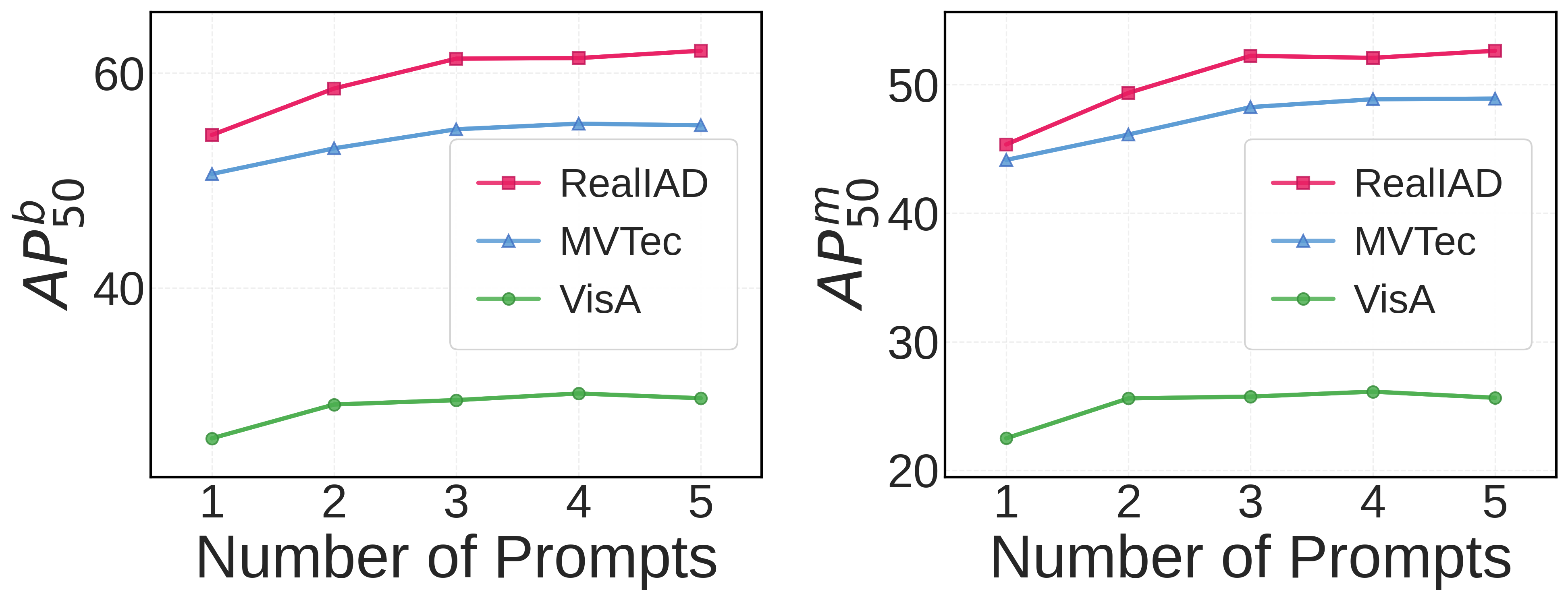}
 \end{center}
 \caption{Effect of the number of prompt samples per defect class.}
\label{fig:prompt_number}
\end{figure}

\noindent\textbf{Query Selection Ablation}  
Comparisons with fixed learnable queries and heuristic top-\textit{K} similarity-based selection (Tab.~\ref{tab:ablation_PQS_multi}) show simply replacing learnable queries with prompt-relevant tokens yields only marginal improvements. In contrast, PQS achieves substantial gains across both datasets and all metrics by learning a differentiable, prompt-conditioned query distribution.

\noindent\textbf{Seen in-domain performance.}
To verify that the proposed framework remains competitive under fully supervised conditions, we evaluate it on Real-IAD using all defect categories for training, with images split only into train and test sets (Tab.~\ref{tab:closed_result}). This setup corresponds to a pure closed-set evaluation. Notably, DINOv and T-Rex2 fail significantly, struggling to recognize even seen classes. This confirms that naive prompt-to-region matching leads to prompt embedding collapse when faced with the high intra-class variance of industrial defects. In contrast, our model achieves performance on par with strong closed-set baselines. These results indicate that \textit{UniSpector} generalizes well to open-set scenarios while preserving high accuracy on seen sets.

\noindent\textbf{Visualization of Prompt Embedding}  
We visualize the prompt embeddings learned by \textit{UniSpector} and DINOv in a 3D embedding space. As shown in Fig.~\ref{fig:prompt_embeddings}, \textit{UniSpector} forms compact clusters within each defect class and clear angular separation across classes, indicating substantial reductions in both intra-class variance and inter-class separability. These improvements arise from two factors: (1) the SSPE, which injects orientation-invariant spectral cues that mitigate rotation-induced intra-variance and enrich feature representation beyond pixel-level, and (2) the CPE, which explicitly enforces intra-class compactness and inter-class angular margins.  
In contrast, DINOv produces scattered and less separable embeddings, reflecting weak intra-class grouping—a consequence of relying solely on prompt-to-region matching without imposing a coherent semantic structure.  
Overall, \textit{UniSpector} yield a semantically organized and transferable embedding space, where unseen sets (e.g., \textit{Missing parts on U block}) naturally position near related seen sets (e.g., \textit{Abrasion on SIM card set}), enabling stronger open-set generalization.

\noindent\textbf{Intra-class Variance Analysis}  
To quantitatively assess the compactness of prompt embeddings, we measure the average cosine similarity between embeddings of instances within the same class. As shown in Fig.~\ref{fig:intra-cosine}, \textit{UniSpector} achieves higher intra-class similarity for seen sets while exhibiting only slightly reduced similarity for unseen sets. In contrast, T-Rex2 and DINOv show lower intra-class similarity for seen sets and more pronounced degradation for unseen sets. These results quantitatively confirm that \textit{UniSpector} learns a more generalized embedding space, supporting improved open-set transferability.

\noindent\textbf{Impact of prompt samples.} We further investigate how the number of prompt per defect class influences performance on MVTec, VisA and RealIAD. Performance stabilizes with only 1--3 prompt images per class (Fig.~\ref{fig:prompt_number}). This data efficiency suggests the learned space captures core defect semantics without requiring extensive exemplar diversity, making it highly practical for industrial settings, where high-quality exemplar collection is often limited.
\section{Conclusion}
In this paper, we introduced \textit{UniSpector}, a universal visual prompting framework that addresses embedding collapse in open-set industrial inspection. To resolve this, we proposed the SSPE to extract refined, orientation-invariant representations, providing a solid basis for the CPE to explicitly regularize the prompt space into a semantically structured angular topology. This topological structuring ensures intra-class compactness and inter-class separation, enabling robust generalization to novel defects.
Beyond its state-of-the-art performance, \textit{UniSpector} offers critical insights into generic visual prompting while establishing a scalable, retraining-free foundation for next-generation industrial inspection.

\newpage
\appendix

\twocolumn[
  \begin{center}
    \textbf{\Large \textit{UniSpector}: Towards Universal Open-set Defect Recognition via Spectral-Contrastive Visual Prompting\\Appendix}
    \vspace{20pt}
  \end{center}
]

\section{Dataset Details}
\hypertarget{supp:section1}{}
We construct the \textit{Inspect Anything} (InsA) benchmark from seven industrial inspection datasets: GC10-DET~\cite{lv2020deep}, Magnetic Tile Surface Defect~\cite{huang2020surface}, Real-IAD~\cite{wang2024real}, MVTec AD~\cite{bergmann2019mvtec}, 3CAD~\cite{yang20253cad}, VISION~\cite{bai2023vision}, and VisA~\cite{zou2022spot}. 
Among these, GC10-DET, Magnetic Tile Surface Defect, Real-IAD, and MVTec AD are used as \emph{in-domain} datasets, with defect categories partitioned into \emph{seen} classes for training and \emph{unseen} classes for testing. In contrast, 3CAD, VISION, and VisA serve as \emph{cross-domain} datasets, used exclusively for evaluating transferability to novel defect types and object appearances. 
To ensure robust estimates of generalization performance, we form three independent seen–unseen splits with random seeds $\{42,82,777\}$, holding out roughly 25\% of defect categories in each in-domain dataset as \emph{unseen}. 

\subsection{Dataset Composition and Characteristics}
Table~\ref{tab:dataset_summary} summarizes key characteristics of each dataset, including product types, materials, and HSV color statistics. 
The InsA benchmark is composed of diverse products and materials, enabling evaluation of open-set visual inspection performance across varied distributions. 
Defect images are captured under a mix of distinct imaging conditions (e.g., very bright or very dark), resulting in generally large standard deviations in HSV channels. 
These distributions also vary across datasets, leading to differing mean values and further highlighting the challenge of generalizing visual inspection models across heterogeneous industrial data.

% \noindent\textbf{GC10.}
\noindent\textbf{GC10~\cite{lv2020deep}.}
GC10-DET is a defect detection dataset collected from metallic surfaces in industrial environments. It comprises roughly 2,300 defect images annotated with bounding boxes for ten defect categories: \emph{silk spot}, \emph{welding line}, \emph{punching hole}, \emph{water spot}, \emph{crescent gap}, \emph{oil spot}, \emph{inclusion}, \emph{waist folding}, \emph{crease}, and \emph{rolled pit}. For each random seed, we select three categories as \emph{in-domain unseen} test set and use the remaining seven categories as \emph{in-domain seen} training set under that seed. 

\noindent\textbf{MagneticTile~\cite{huang2020surface}.}
The magnetic tile surface defect dataset contains images of magnetic tile surfaces collected from industrial production lines, covering both defective and defect-free samples.
It provides pixel-level annotations for five defect categories—\emph{blowhole}, \emph{crack}, \emph{break}, \emph{fray}, and \emph{uneven}—while normal images are discarded in our benchmark to focus solely on faulty cases. We retain approximately 400 defect images from the original release. For each random seed, we select one defect category as an \emph{in-domain unseen} test set and use the remaining four categories as \emph{in-domain seen} classes for training under that seed.

\begin{table*}[t]
  \small
  \centering
  \setlength{\tabcolsep}{2mm}{
  \begin{tabular}{c|c|c|ccc}
    \toprule
    \multirow{2}{*}{Dataset} & 
    \multirow{2}{*}{Product} & 
    \multirow{2}{*}{Material} & 
    \multicolumn{3}{c}{Color Distribution Statistics} \\
    & & & H & S & V \\
    \midrule
    GC10~\cite{lv2020deep} 
      & Steel & Steel 
      & $0.0 \pm 0.0$ & $0.0 \pm 0.0$ & $85.7 \pm 39.6$ \\
    MagneticTile~\cite{huang2020surface} 
      & Magnetic Tile & Steel 
      & $0.0 \pm 0.0$ & $0.0 \pm 0.0$ & $109.4 \pm 47.4$ \\
    Real-IAD~\cite{wang2024real} 
      & PCB, Toy Brick, Transistor, etc. 
      & Plastic, Rubber, Wood, etc. 
      & $33.4 \pm 49.5$ & $40.6 \pm 65.7$ & $110.2 \pm 105.5$ \\
    MVTec AD~\cite{bergmann2019mvtec} 
      & Cable, Hazelnut, Tile, etc. 
      & Glass, Metal, Fabric, etc. 
      & $46.7 \pm 55.1$ & $46.0 \pm 47.8$ & $120.5 \pm 66.7$ \\
    \midrule
    3CAD~\cite{yang20253cad} 
      & Camera Cover, Tablet PC, etc. 
      & Aluminum, Copper, etc. 
      & $7.2 \pm 19.2$ & $16.7 \pm 49.0$ & $78.6 \pm 70.5$ \\
    VISION~\cite{bai2023vision} 
      & Capacitor, Lens, Screw, etc. 
      & Plastic, Steel, Wood, etc. 
      & $25.2 \pm 41.1$ & $38.4 \pm 70.7$ & $106.3 \pm 83.3$ \\
    VisA~\cite{zou2022spot} 
      & Candle, Capsule, Macaroni, etc. 
      & Plastic, Food, etc. 
      & $47.7 \pm 36.4$ & $95.7 \pm 71.9$ & $101.9 \pm 66.1$ \\
    \bottomrule
  \end{tabular}}
  \caption{Dataset statistics used in the InsA benchmark, including their product types, materials, and the color distribution statistics (mean~$\pm$~std) for each HSV channel.}
  \label{tab:dataset_summary}
\end{table*}

\noindent\textbf{Real-IAD~\cite{wang2024real}.}
Real-IAD is a multi-view industrial anomaly dataset comprising \emph{30} real-world objects fabricated from diverse materials (e.g., plastic, rubber). Each object is captured under five viewpoints and exhibits \emph{2–5} distinct defect modes selected from eight defect families (i.e., \emph{pit}, \emph{deformation}, \emph{abrasion}, \emph{scratch}, \emph{damage}, \emph{missing parts}, \emph{foreign objects}, and \emph{contamination}). Only viewpoints equipped with pixel-level defect masks are retained as anomaly samples, and we use the official $1024 \times 1024$-resolution version of the images. From the binary defect masks, we perform connected-component labeling with 8-connectivity to obtain instance-level segments, and discard components whose width or height is smaller than $1\%$ of the image width or height to remove tiny noisy polygons. For each split, we select \emph{28} defect categories as \emph{in-domain unseen} test classes and use the remaining categories as \emph{in-domain seen} classes for training.

\noindent\textbf{MVTec AD~\cite{bergmann2019mvtec}.}
MVTec AD is a real-world industrial anomaly dataset that contains defective images across multiple object and texture categories, with various defect types such as contamination, oil, cuts, and cracks. In our benchmark, we discard all ``good'' images and use only anomalous samples with polygon masks. From these masks, we extract individual defect instances and remove components whose width or height is smaller than $1\%$ of the corresponding image dimension, thereby filtering out tiny noisy regions.

\noindent\textbf{3CAD~\cite{yang20253cad}.}
3CAD is a large-scale anomaly detection dataset collected from real 3C product manufacturing lines, covering representative defects that arise in practical production environments. It focuses on parts made of three common materials (Aluminum, Iron, and Copper) and includes multiple types of 3C components (e.g., camera covers, tablets, and PCs). Since 3CAD is designed for anomaly detection, we discard the ``good'' class without polygon annotations and also remove the \textit{Multiple-defects} class, whose defect instances cannot be assigned to specific categories. After this filtering, we obtain a total of \emph{46} defect categories. From the remaining polygons, we extract individual defect instances and discard components whose width or height is smaller than $1\%$ of the corresponding image dimension to eliminate tiny noisy regions.

\noindent\textbf{VISION~\cite{bai2023vision}.}
The VISION benchmark unifies \emph{14} industrial inspection subsets, each corresponding to a distinct object class from real manufacturing lines and captured at its native (often high) resolution; consequently, image size varies across subsets. The corpus provides pixel-level instance masks for \emph{44} defect categories. For InsA, we retain only images in the \textit{train} and \textit{val} partitions that include polygon annotations and discard the \textit{inference} split, whose labels are withheld. From the polygon masks, we remove segments whose width or height is smaller than $1\%$ of the width or height of the image to filter out tiny noisy regions. All 44 defect categories are used exclusively for cross-domain evaluation, and none of them are included in the training data.

\noindent\textbf{VisA~\cite{zou2022spot}.}
VisA is an industrial visual inspection dataset containing normal and defective images from \emph{12} object categories, some of which exhibit large variations in object location and pose across images. It covers both surface-level defects (e.g., scratches, dents) and structural defects (e.g., misplacement). In our benchmark, we discard all normal images and retain only those that contain at least one annotated defective region. We also exclude the \textit{Other} defect class, whose semantics are not clearly specified. From the binary defect masks, we apply connected-component labeling with 8-connectivity to obtain instance-level segments, and discard components whose width or height is smaller than $1\%$ of the corresponding image dimension, thereby filtering out tiny noisy regions.

\begin{figure*}[!t]
\begin{center}
 \includegraphics[width=1.0\linewidth,height=7cm]{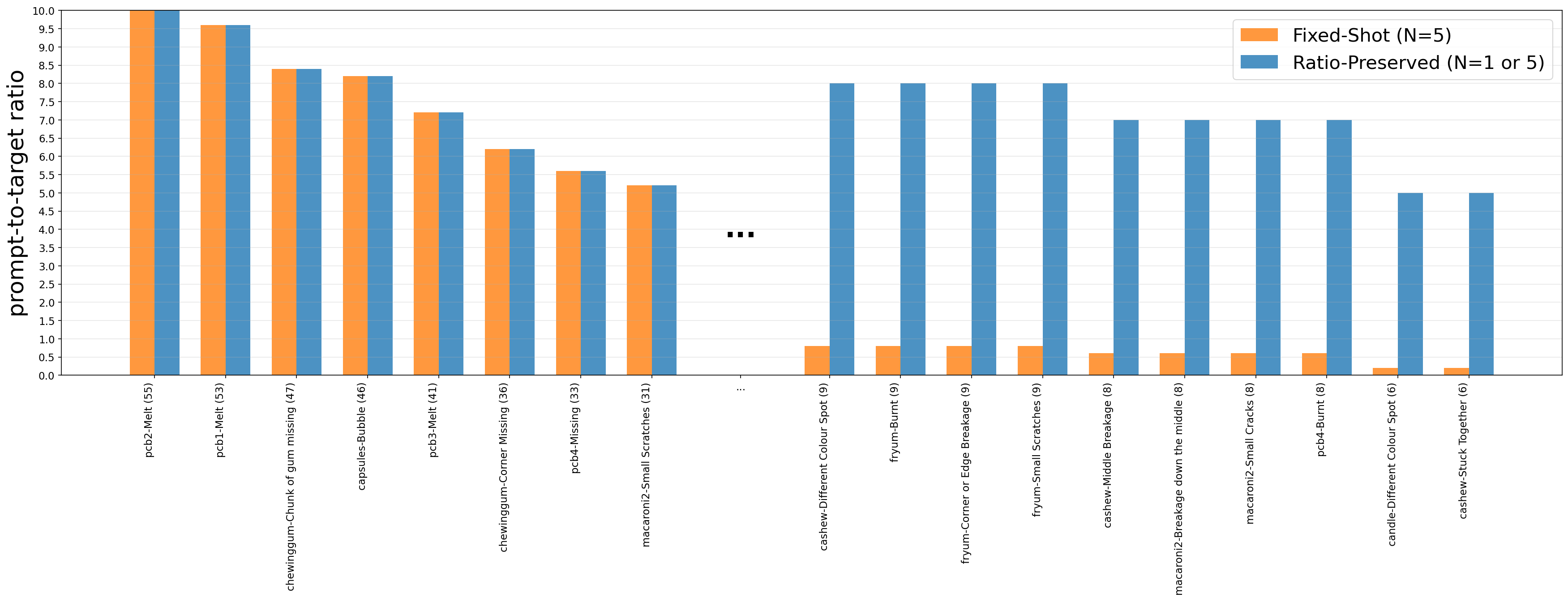}
 \end{center}
 \caption{Distribution of Prompt-to-Target Ratios across Defect Classes. It illustrates the prompt-to-target ratio for each defect class (with sample sizes). We compare our adaptive prompt strategy (Blue) against the fixed prompt strategy (Orange). By reducing the number of prompts to one for tail classes (total images $\le 10$), the adaptive strategy prevents an irrational ratio in the tail classes, ensuring a more balanced and fair evaluation across the entire InsA dataset.}
\label{fig:prompt_ratio}
\end{figure*}

\subsection{Prompt Allocation}
\noindent\textbf{Selection of Prompt data}
To ensure class purity, prompt samples are selected only from \emph{single-label} samples that contain the target defect alone; mixing co-occurring other categories into a prompt would inject ambiguous cues and dilute the class‐specific signal. 

\noindent\textbf{Image-level vs. Instance-level}
Furthermore, unlike conventional few-shot protocols~\cite{kang2019few,yan2019meta,wang2020frustratingly} that count shots per instance (box or mask), we define \emph{k} with respect to entire \emph{images}. In industrial defect inspection a single photograph often exhibits multiple defect instances whose visual similarity renders selective prompting ill-posed; excluding certain regions while prompting others would implicitly guide the detector to ignore genuine defects within the same frame. Counting shots at the image level therefore mirrors real deployment conditions and yields a fairer, more interpretable comparison across categories and methods.

\noindent\textbf{Prompt Size}
For each defect category, we default to 5 visual prompt images during evaluation. However, to ensure a balanced evaluation across all dataset in InsA. For the "tail classes" (total images ≤ 10 ), we restrict the number of prompts to 1, as assigning 5 prompts to classes with very few target images leads to an irrational prompt-to-target ratio. As shown in Figure~\ref{fig:prompt_ratio}, by reducing the prompt to a single image for these cases, we achieved a more balanced and reasonable evaluation setting for long-tailed defect distributions. Notably, potential selection bias from using a single prompt is mitigated by averaging results across multiple random seeds, ensuring a robust evaluation.

\begin{figure*}[!t]
\begin{center}
 \includegraphics[width=1.0\linewidth,height=10cm]{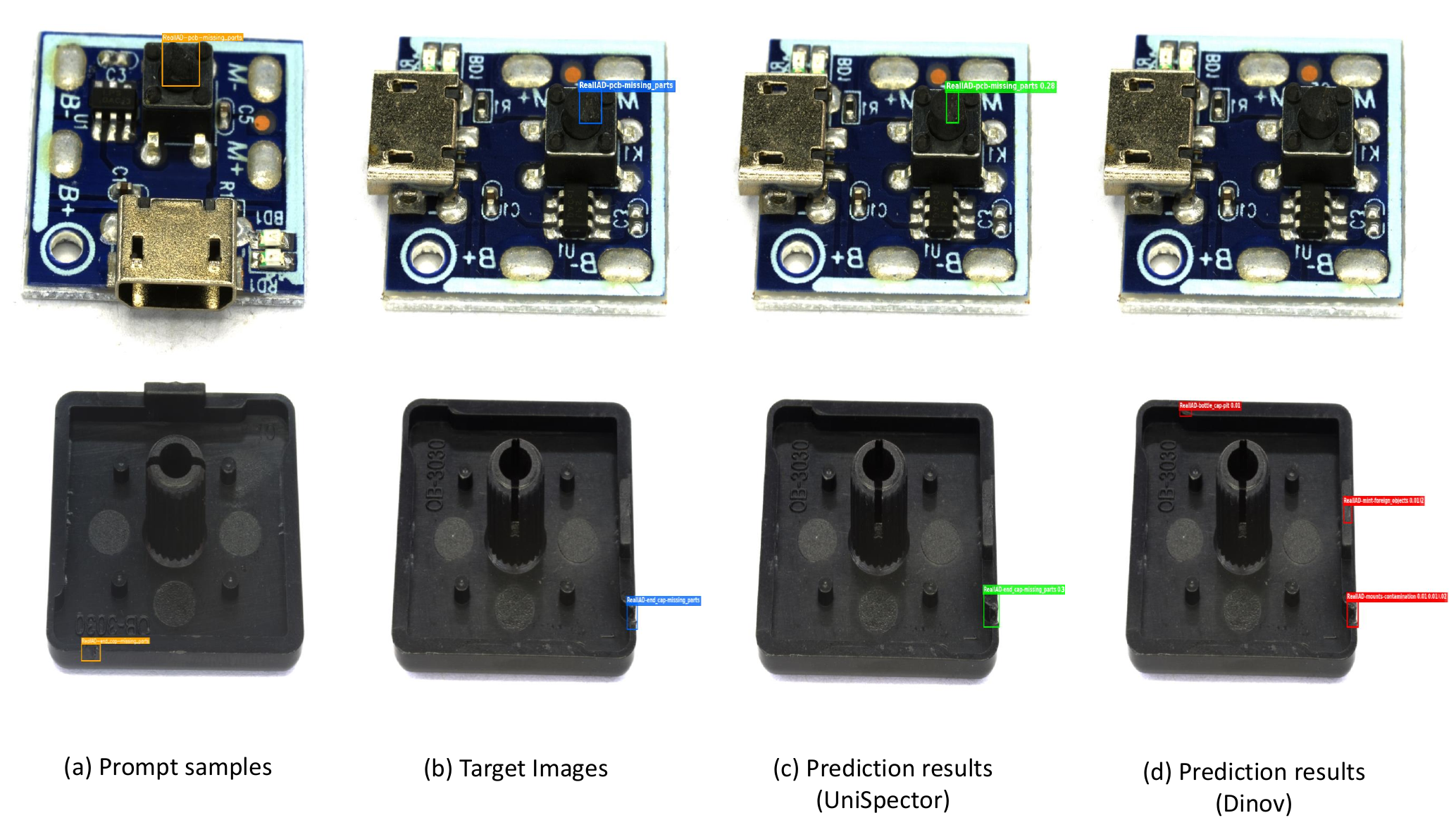}
 \end{center}
 \caption{\textit{UniSpector} is capable of recognizing unseen defects via visual prompts. 
(a) Orange box: user-specified prompt region. 
(b) Blue box: corresponding ground-truth in the target image. 
(c) Green boxes: correct predictions by \textit{UniSpector}, accurately localizing subtle defects. 
(d) Red boxes: DINOv predictions, showing failure to localize the prompted defect.}
\label{fig:pred_figure}
\end{figure*}

\begin{algorithm}
\caption{Radial Frequency Feature Extraction}
\label{alg:radial_freq_compact}
\begin{algorithmic}[1]
\REQUIRE Image $I$, Bounding boxes $\mathcal{B}$, Frequency bins $J$
\ENSURE Spectral features $\mathbf{F} \in \mathbb{R}^{K \times J}$
\FOR{each bounding box $b_i \in \mathcal{B}$} 
  \STATE $I_{\text{crop}} \leftarrow \text{Crop}(I, b_i)$ and convert to grayscale
  \STATE $M \leftarrow |\text{FFTShift}(\text{FFT2D}(I_{\text{crop}}))|$
  \STATE $R \leftarrow \sqrt{(X - c_x)^2 + (Y - c_y)^2}$
  \STATE $R_{\text{norm}} \leftarrow R / \max(R)$
  \FOR{$j = 1$ to $J$}
    \STATE $f_j \leftarrow \text{mean}(M[R_{\text{norm}} \in \text{bin}_j])$
  \ENDFOR
  \STATE $\mathbf{f}_i \leftarrow \text{L2Normalize}([f_1, ..., f_J])$
\ENDFOR
\RETURN $\mathbf{F} = [\mathbf{f}_1, ..., \mathbf{f}_K]^T$
\end{algorithmic}
\end{algorithm}

\hypertarget{supp:section2}{}
\section{Details of Radial Frequency Extraction}
We present the radial frequency extraction procedure in Algorithm~\ref{alg:radial_freq_compact}. Given a region crop, 2D FFT is computed and the resulting frequency magnitudes are aggregated into radial bins and L2-normalized. This process introduces minimal computational overhead while providing complementary texture cues to the spatial features.

\section{Implementation Details}
\hypertarget{supp:section31}{}
\subsection{Architectural Details}
\textit{UniSpector} is built upon the MaskDINO~\cite{li2023mask} architecture, an encoder-decoder framework designed for unified object detection and segmentation. For the encoder, we employ a Swin-T backbone with hierarchical stages to extract multi-scale feature maps. For fair comparison, the same SwinT backbone is used for DINOv, T-Rex2 and \textit{UniSpector}, while YOLOE utilize its native YOLO11m~\cite{yolo11_ultralytics} backbone due to architectural constraints. The decoder is composed of 9 Deformable Transformer layers with 8 attention heads. Each layer consists of self-attention, multi-scale deformable cross-attention, and an FFN. 
\hypertarget{supp:section32}{}
\subsection{Training Objective}
We follow the standard MaskDINO training recipe, incorporating denoising (DN) training and auxiliary deep supervision across all $L$ decoder layers to accelerate convergence. Bounding box regression is supervised by $L_1$ and GIoU losses, while mask segmentation employs a combination of BCE and Dice losses. Classification is addressed by a Focal loss measuring the similarity between object queries and class prototypes. The bipartite matching between predictions and ground truths is established via Hungarian matching. Finally, the proposed $\mathcal{L}_{\text{CPE}}$ is integrated with these standard terms to form total objective:
\begin{equation}
\begin{aligned}
    \mathcal{L}_{\text{total}} = \sum_{l=0}^{L-1} \Big( & \lambda_{\text{cls}}\mathcal{L}_{\text{cls}}^{(l)} + \lambda_{\text{L1}}\mathcal{L}_{\text{L1}}^{(l)} + \lambda_{\text{giou}}\mathcal{L}_{\text{giou}}^{(l)} +\lambda_{\text{bce}}\mathcal{L}_{\text{bce}}^{(l)} \\ & + \lambda_{\text{dice}}\mathcal{L}_{\text{dice}}^{(l)}\Big) + \mathcal{L}_{\text{DN}}  + \lambda_{\text{CPE}}\mathcal{L}_{\text{CPE}}
\end{aligned}
\end{equation}
where each $\lambda$ denotes the corresponding loss weight. Note that $\mathcal{L}_{\text{DN}}$ is the denoising counterpart of the same base losses with the same coefficients as their corresponding main losses. Under this joint optimization, we use the default weights: $\lambda_{\text{cls}}=4$, $\lambda_{\text{L1}}=5$, $\lambda_{\text{giou}}=2$, $\lambda_{\text{bce}}=5$, $\lambda_{\text{dice}}=5$, and $\lambda_{\text{CPE}}=1$.
\hypertarget{supp:section33}{}
\subsection{Details of Inference Phase}
\begin{figure}
\begin{center}
 \includegraphics[width=0.9\linewidth,height=6cm]{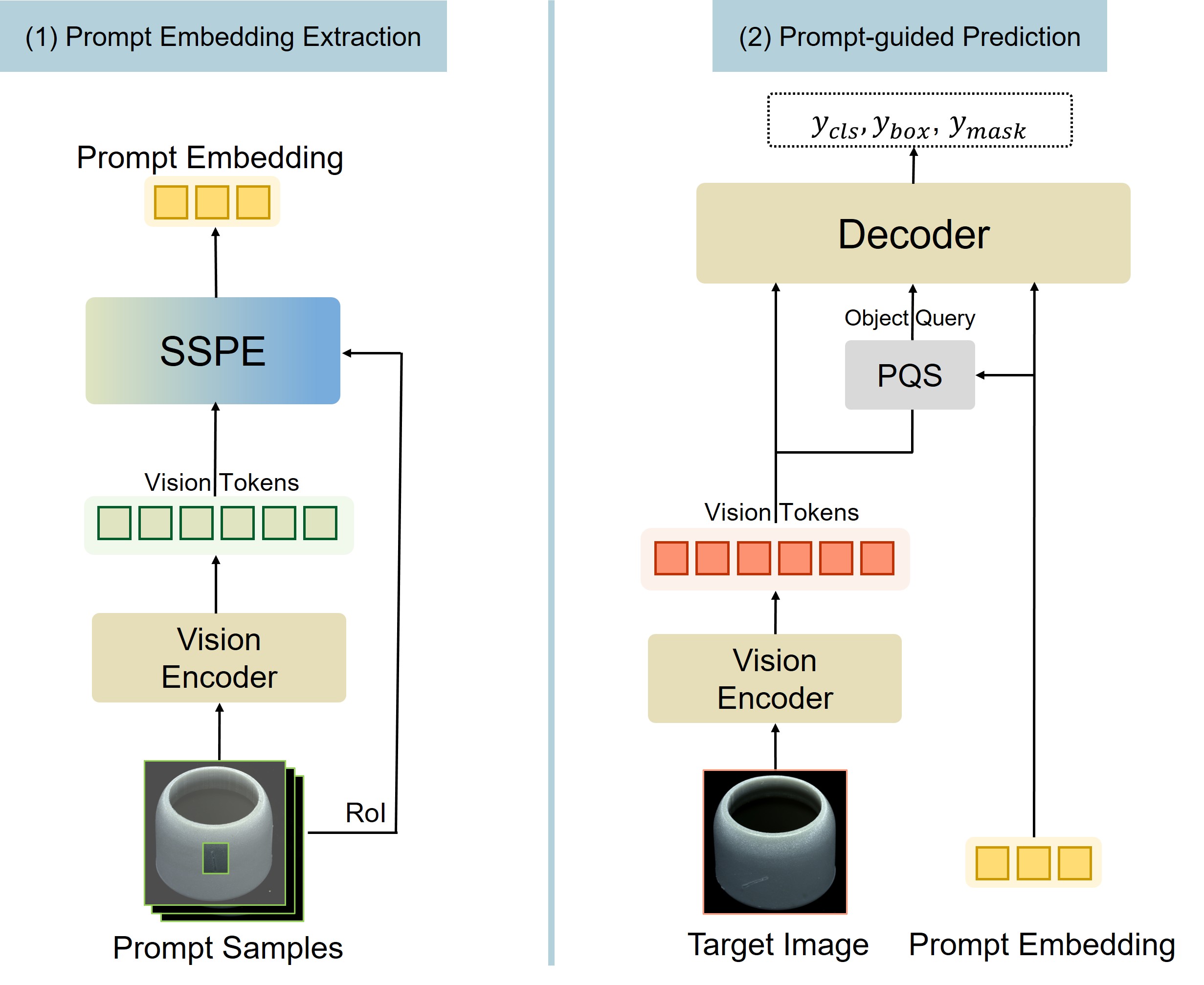}
 \end{center}
 \caption{Detailed view of the inference phase. Unlike the training phase (see Fig.~\ref{fig:architecture} in the main manuscript), inference is decoupled into ``Prompt Embedding Extraction" and ``Prompt-guided Prediction" to minimize computational overhead by leveraging cached embeddings.}
\label{fig:inference_detail}
\end{figure}

As illustrated in Figure~\ref{fig:inference_detail}, \textit{UniSpector} decouples the inference process into two distinct stages to minimize computational overhead. Specifically, Prompt Embedding Extraction is performed as a one-time offline process to extract prompt embeddings. During the subsequent Prompt-guided Prediction stage, these stored embeddings are loaded directly, bypassing redundant re-extraction of prompt embeddings for every query. Notably, this architecture allows \textit{UniSpector} to seamlessly integrate prompt embeddings for any defect category of interest, enabling flexible and efficient deployment in open-set scenarios.

\hypertarget{supp:section34}{}
\subsection{Training settings}
All Swin-T-based models are initialized with the official DINOv pre-trained wights, while YOLOE is initialized from its publicly available checkpoint. Training is conducted for 20,000 steps with a total batch size of 48 on two NVIDIA H100 GPUs. We use the AdamW optimizer with a linear learning rate schedule, a base learning rate of \(1 \times 10^{-4}\), a 10-step warm-up, and a weight decay of 0.05. Data augmentation includes random horizontal flipping, contrast adjustment (0.8--1.1), and brightness adjustment (0.5--1.3). All models are trained at an input resolution of \(720 \times 720\) pixels. The radial frequency distribution is computed using 256 bins. The scaling factor is set to \(s = 30.0\) and the angular margin to \(m = 0.5\), which empirically stabilizes the optimization. We use \(N_q=300\) object queries and set the Gumbel-Softmax temperature to \(\tau=1.0\).

\section{Additional analysis}
\subsection{Robustness to the quality of prompts}
To analyze robustness to prompt quality which is indeed critical factor for practical deployment. We conducted additional experiments simulating realistic industrial scenarios along two dimensions: image and annotation.
\begin{figure}[h]
    \centering
    \includegraphics[width=1.0  \columnwidth]{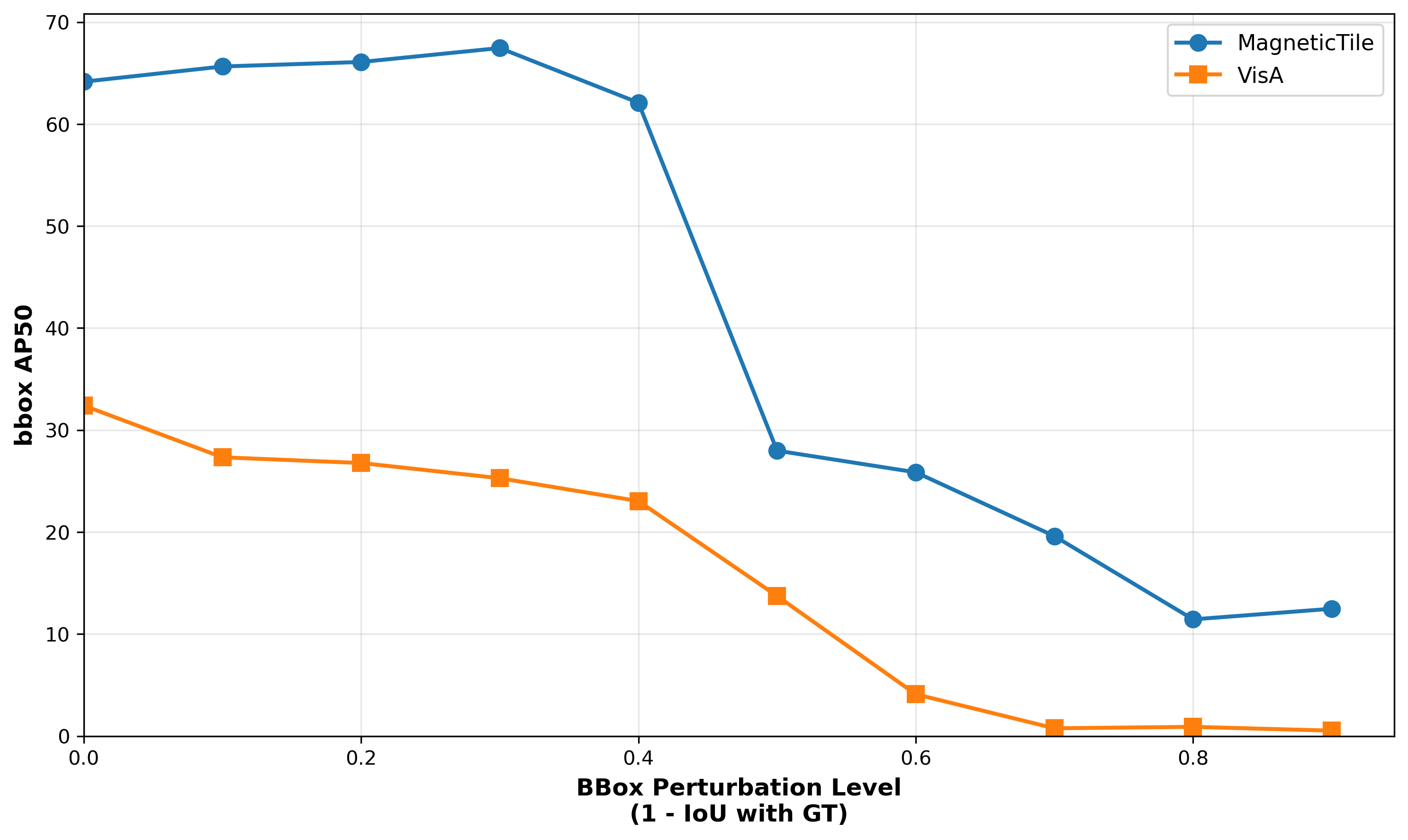}
    \caption{Robustness against prompt annotation (averaged over 10 trials per IoU). Each perturbed box is constrained to a specific IoU level relative to the ground-truth label.}
    \label{fig:annotation}
\end{figure}

\begin{table}[h]
\centering
\caption{Experimental analysis under Gaussian Blur on the prompt images. We evaluate the impact of varying blur levels ($\sigma$) on the performance.}
\label{tab:blur}
\setlength{\tabcolsep}{2pt}
\begin{tabular}{ll|ccccccc}
\toprule
Dataset & Metric & Clean & $0.5\sigma$ & $1.0\sigma$ & $1.5\sigma$ & $2.0\sigma$ & $2.5\sigma$ & $3.0\sigma$ \\ 
\midrule
VisA & \scalebox{0.7}{$AP^b_{50}$} & 32.5 & 33.3 & 32.0 & 30.0 & 27.4 & 25.8 & 23.4 \\
 & \scalebox{0.7}{$AP^m_{50}$} & 28.2 & 28.3 & 27.6 & 26.3 & 23.8 & 22.3 & 20.4 \\
MTile & \scalebox{0.7}{$AP^b_{50}$} & 64.2 & 64.9 & 64.1 & 63.5 & 62.0 & 60.1 & 57.0 \\
 & \scalebox{0.7}{$AP^m_{50}$} & 38.9 & 39.1 & 38.8 & 37.8 & 36.4 & 34.3 & 31.0 \\ 
\bottomrule
\end{tabular}
\end{table}

\noindent\textbf{Image Degradation}
To simulate blurry conditions, we evaluated the performance under Gaussian blur kernels with controlled standard deviations ($\sigma$) on the prompt images. The results are summarized in Table~\ref{tab:blur}. Interestingly, a slight performance improvement was observed at $0.5\sigma$ blur. This improvement likely stems from the suppression of high-frequency noise, yielding prompt embeddings that are less biased toward instance-specific artifacts. The performance differences remain small up to $1.5\sigma$, while a clear degradation is observed beyond $2.0\sigma$, where the image quality degradation becomes visually noticeable.
Performance is stable under Gaussian blur up to $\sigma=1.5$ (Tab.~\ref{tab:blur}) and box perturbations down to IoU 0.3 (Fig.~\ref{fig:annotation}, proving robustness.

\noindent\textbf{Noisy Annotation}
We further examined robustness by randomly perturbing the prompt bounding boxes to achieve IoU levels ranging from 0.1 to 0.9 relative to the ground truth (GT). Each configuration was averaged over 10 independent trials. As shown in Figure~\ref{fig:annotation}, UniSpector maintains high performance down to an IoU of 0.3, demonstrating significant tolerance to imprecise annotations. Beyond this threshold, the perturbed boxes begin to encompass background or non-defect regions rather than the target defect, leading to an expected drop in precision as the prompt's semantic signal becomes corrupted.

\subsection{Analysis of In-Context Learning Approaches}
We conducted quantitative comparisons with representative in-context learning methods, PerSAM[1] and VRP-SAM[2], as shown in the table below. When evaluated in a training-free manner, PerSAM achieved an average $\mathrm{AP}^{m}_{50}$ of 6.0 due to the significant domain gap. The fine-tuning variant (PerSAM-F) yielded only a marginal improvement of +1.4 in average performance. We attribute this to a representation bottleneck: both PerSAM and VRP-SAM fundamentally rely on the representational capacity of frozen SAM encoders. Relying on frozen SAM general-object representations, PerSAM and VRP-SAM fail to generalize to InsA even after fine-tuning. Because the feature space of SAM, which is primarily trained on general real-world objects, may not be optimally suited for discriminating fine-grained features of subtle and complex defects. For VRP-SAM, since the authors do not release pre-trained weights for the VRP-Encoder, we report only the performance after training the VRP-Encoder on our in-domain seen set.
Relying on frozen SAM general-object representations, PerSAM and VRP-SAM fail to generalize to InsA even after fine-tuning (Tab.~\ref{tab:icl_comparison}). 
\begin{table}[h]
\centering
\caption{Results of in-context learning methods ($AP^m_{50}$) on InsA benchmark. $^\dagger$ indicates results obtained by fine-tuning on the in-domain seen split of the proposed protocol. }
\label{tab:icl_comparison}
\scriptsize
\setlength{\tabcolsep}{2pt}
\begin{tabular}{l|ccccccc|c}
\toprule
Method & GC10 & MTile & RIAD & MVTe & 3CAD & VISN & VisA & Avg \\ 
\midrule
PerSAM & 0.0 & 33.0 & 0.8 & 4.8 & 0.1 & 0.3 & 3.0 & 6.0 \\
PerSAM$^\dagger$ & 0.1 & 34.2 & 1.5 & 8.7 & 0.2 & 1.0 & 6.0 & 7.4 \\
VRP-SAM$^\dagger$ & 0.4 & 12.5 & 22.3 & 18.8 & 3.2 & 4.7 & 12.2 & 10.6 \\
\bottomrule
\end{tabular}
\end{table}

\subsection{Visualization}
Figure~\ref{fig:pred_figure} presents qualitative results for open-set defect detection.
Given a user-specified region in the prompt images (orange bounding box), \textit{UniSpector} successfully identifies corresponding defect instances in the target image. As illustrated in Figure~\ref{fig:pred_figure}(c), \textit{UniSpector} accurately captures even subtle and fine-grained defects in PCB (TOP). Moreover, despite the pronounced rotational difference between the prompt and target defect instance, \textit{UniSpector} consistently localizes the prompted defect type (BTM).

{
    \small
    \bibliographystyle{ieeenat_fullname}
    \bibliography{main}
}


\end{document}